\documentclass{article}

\usepackage{arxiv}

\usepackage[utf8]{inputenc} % allow utf-8 input
\usepackage[T1]{fontenc}    % use 8-bit T1 fonts
\usepackage{hyperref}       % hyperlinks
\usepackage{url}            % simple URL typesetting
\usepackage{booktabs}       % professional-quality tables
\usepackage{amsfonts}       % blackboard math symbols
\usepackage{nicefrac}       % compact symbols for 1/2, etc.
\usepackage{microtype}      % microtypography
\usepackage{lipsum}
\usepackage{graphicx}
\usepackage{wrapfig}
\usepackage{natbib}
\usepackage{amsmath}
\usepackage{algorithmic}
\usepackage{textcomp}
\usepackage{multirow}
\usepackage{makecell}
\usepackage{arydshln}
\usepackage{xr}
\usepackage{cases}
\usepackage{amsthm}
\usepackage{siunitx}
\usepackage{bm}
\usepackage{fancyhdr}

\fancypagestyle{firststyle}
{
   \fancyhead{}
   \fancyfoot{}
   \fancyfoot[L]{\small This is a revised preprint. The Version of Record of this contribution is published in Medical Image Computing and Computer-Assisted Intervention – MICCAI 2025, Lecture Notes in Computer Science, Vol. 15961,, and is available online at \url{https://doi.org/10.1007/978-3-032-04937-7_38}.}
}
\fancypagestyle{emptystyle}
{
   \fancyhead{}
   \fancyfoot{}
}
\DeclareMathOperator{\Var}{Var}

\theoremstyle{plain}

\theoremstyle{definition}

\theoremstyle{remark}

\title{New multimodal similarity measure for image registration via modeling local functional dependence with linear combination of learned basis functions}

\author{
 Joel Honkamaa \\
  Department of Computer Science\\
  Aalto University
   \And
 Pekka Marttinen \\
  Department of Computer Science\\
  Aalto University 
}

\begin{document}
\maketitle
\begin{abstract}
The deformable registration of images of different modalities, essential in many medical imaging applications, remains challenging. The main challenge is developing a robust measure for image overlap despite the compared images capturing different aspects of the underlying tissue. Here, we explore similarity metrics based on functional dependence between intensity values of registered images. Although functional dependence is too restrictive on the global scale, earlier work has shown competitive performance in deformable registration when such measures are applied over small enough contexts. We confirm this finding and further develop the idea by modeling local functional dependence via the linear basis function model with the basis functions learned jointly with the deformation. The measure can be implemented via convolutions, making it efficient to compute on GPUs. We release the method as an easy-to-use tool and show good performance on three datasets compared to well-established baseline and earlier functional dependence-based methods.

% Authors must provide keywords and are not allowed to remove this Keyword section.

\end{abstract}
\thispagestyle{firststyle}
\section{Introduction}
The multimodal medical image registration aims to find a mapping between the anatomical coordinates of images of different modalities. This is an important prerequisite for the efficient utilization of complementary information provided by different imaging modalities. In deformable registration, the mappings are not limited to linear (affine). The problem is difficult and, while studied intensively for a few decades, it remains an active topic of research.

The registration of distorted images is traditionally formulated as an optimization problem that involves a similarity term and a regularization term \cite{oliveira2014medical}. In multimodal image registration, measuring similarity is particularly difficult due to the complex dependency between the registered images.
\begin{figure}[t]
\includegraphics[width=\textwidth]{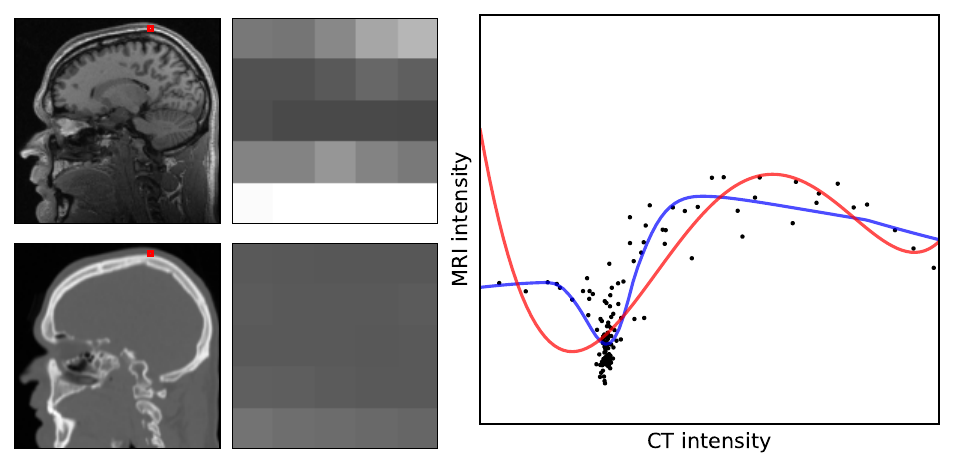}
\caption{We measure multimodal similarity via residuals of locally fitted functions (over sliding window). Each point on the right describes intensity value pair for voxels at identical locations in the patches (only one slice of the 3D volumes is shown). The points are weighted by the distance from the patch center. \textbf{Blue curve:} Learning basis (globally) allows reasonably good fit with very few basis functions (number of terms $J=5$). \textbf{Red curve:} Polynomial functions (number of terms $J=6$) struggle to fit the high frequencies. \copyright Copyright CERMEP – Imagerie du vivant, www.cermep.fr and Hospices Civils de Lyon. All rights reserved.} \label{fig:learned_basis}
\end{figure}
Successful methods include measuring similarity with mutual information (MI) \cite{viola1997alignment,pluim2003mutual}, a measure of statistical dependence. Another method of multimodal (affine) registration is the correlation ratio (CR) \cite{roche1998correlation}, which measures the more restrictive functional dependence between intensity values, addressing the fact that mutual information largely ignores the proximity of intensity values. In addition, CR is efficient to compute. However, functional dependence is, in general, too restrictive, especially for deformable registration.

Both MI and CR are global measures, making them sensitive to spatial changes in statistical relationships between the image intensities, e.g. due to non-uniform bias fields \cite{pluim2003mutual,heinrich2012mind,rivaz2014automatic}. These measures have been defined in local contexts \cite{pluim2003mutual,klein2008automatic,rivaz2014automatic} for which CR is particularly attractive because it is cheaper to localize and functional dependency is often sufficient for small contexts. In this work, we build on this and develop a well-working image registration based on estimating local functional dependence with the following novel contributions.

\begin{itemize}
    \item We model the intensity relationship with the linear basis function model. The coefficients are fitted locally in closed form, and the basis functions are learned jointly with the deformation. Earlier metrics measuring local functional dependence (RaPTOR \cite{rivaz2014automatic} and SRWCR \cite{gong2018nonrigid}) estimate the intensity relationship using Parzen windowing approximations of conditional expectation (Eq. \ref{eq:correlation_ratio_conditional_expectation}). RaPTOR uses a smooth variant of Eq. \ref{eq:isoset_approximation} whereas SRWCR estimates full joint distribution for all subregions, which is very costly.
    \item We note that the measure can be implemented via a convolution operation, allowing for an easy but efficient GPU implementation.
    \item We release the method as an easy-to-use tool named \textit{Locor} (available at \href{https://github.com/honkamj/locor}{https://github.com/honkamj/locor}) and show good performance on three datasets.
\end{itemize}

\section{Notation}

We view the images to be registered as mappings from physical coordinates to intensity values $I_A: \mathbb{R}^n\to\mathbb{R}$ and $I_B: \mathbb{R}^n\to\mathbb{R}$ where $n$ is the dimensionality of the image. Trilinear interpolation is used to make the images continuous in practice. We limit the mathematical analysis to single channel images for clarity. The task is then to find the deformation $d: \mathbb{R}^n\to \mathbb{R}^n$, describing the mapping from the anatomical locations of the image $I_A$ to the corresponding locations in the image $I_B$. Only part of the domain $\mathbb{R}^n$ contains a valid image, but for clarity, we do not explicitly mention this in the formulas. In general, any comparison measures are applied over the overlapping valid region between the images.

When taking the expected value or variance of images, we interpret the images as random variables on the intensity values (generated by uniformly sampling a coordinate), as is common in the literature \cite{viola1997alignment,roche2000unifying}.

\section{Related work}\label{sec:background}

The correlation ratio (CR) \cite{roche1998correlation} is based on the assumption that there exists a functional relationship between the intensity values of the registered images $f(I_A) \approx I_B$ for some $f: \mathbb{R}\to\mathbb{R}$. The variance of $f(I_A) - I_B$ could then be used as a similarity measure. Since $f$ is unknown a priori, they propose to compute the measure for optimal $f$ (with which the variance is minimized), yielding

\begin{align}
    1 - CR(I_A, I_B) &= \frac{\min_f \Var\left[f(I_A) - I_B\right]}{\Var\left[I_B\right]}
    = \frac{\Var\left[\mathbb{E}\left[I_B | I_A\right] - I_B\right]}{\Var\left[I_B\right]}\label{eq:correlation_ratio_conditional_expectation}.
\end{align}
Assuming that the intensity values are discrete, one can compute the measure exactly as variance of $I_B$ over the isosets (voxels of equal value) of the image $I_A$. Denoting isosets with intensity $a$ as $\Omega_a$, one obtains (here $N:=\sum_a |\Omega_a|$):
\begin{equation}\label{eq:isoset_approximation}
    1 - CR(I_A, I_B) = \frac{\sum_a \sum_{x\in{\Omega_a}} \left(I_B(x) - \frac{1}{|\Omega_a|}\sum_{\tilde{x} \in \Omega_a} I_B(\tilde{x})\right)^2}{N \Var\left[I_B\right]}
\end{equation}
To obtain a differentiable measure, one can use soft windowing to define the isosets (often called Parzen windowing). Alternatively, one could assume a specific parametric form for the function $f$, and in the appendix of \cite{roche2000unifying} and later in \cite{roche2001rigid} Roche et al. explored the function assuming a polynomial form. It should also be noted that the very popular cross-correlation (for intramodality registration) can be derived by assuming an affine form for $f$ \cite{roche2000unifying,avants2008symmetric}.

The correlation ratio, applicable to rigid or affine multimodal registration, was extended to deformable registration in RaPTOR \cite{rivaz2014automatic} by computing the measure over small randomly sampled patches. The idea was further explored in SRWCR \cite{gong2018nonrigid} by Gong et al. where patches are instead sampled over a regularly spaced grid, and voxels within each patch are given less weight further they are from the patch center. The method was implemented for a GPU to make it computationally tractable. Both RaPTOR and SRWCR compute the local CR measure by estimating conditional expectation with Parzen windowing.

\section{Methods}\label{sec:methods}

Maximum likelihood has been used to derive global CR measures in \cite{roche2000unifying}. Here, we first extend the framework to local CR measures by modeling local neighborhoods separately and subsequently aggregate the local losses into a global loss, which we maximize to learn the unknown deformation and other parameters.

Consider a local neighborhood around a point, say $r\in \mathbb{R}^n$, and assume that intensity values of image $I_A$ are modeled with a Gaussian distribution with a mean function $f_r$ that takes as input the intensity values of an (aligned) image $I_B$, with noise variance increasing with the increasing distance from $r$:
\begin{equation}
    I_A(x) \sim \mathcal{N}(f_r(I_B(x)), \phi(x - r)\sigma_r).\label{eq:likelihood}
\end{equation}
In Eq. \ref{eq:likelihood}, $\phi: \mathbb{R}^n \to \mathbb{R}$ is a function that increases with distance from the origin ($\phi(\overline{0}) = 1.0$), and $\sigma_r$ is the variance in $r$. As in \cite{roche2000unifying}, we assume that the errors in the predicted intensity values of $I_A$ at different locations $x$ are independent given $I_B$, and define a likelihood which we denote by $p_r(I_A | I_B, \sigma_r, f_r)$. This likelihood can be maximized to estimate the unknown $f_r$ and $\sigma_r$. Furthermore, the same likelihood will be used to find the (local) deformation from $I_A$ to $I_B$ by finding the deformed $I_B$ with the highest likelihood (combined with a regularization loss), but the deformation is suppressed in the notation for clarity.

The standard formula for the maximized log-likelihood for a Gaussian model (maximization w.r.t. $\sigma_r$), yields a local loss (similar to a weighted least squares):
\begin{equation}
\mathcal{L}_r:=\max_{\sigma_r} \log p_r(I_A | I_B, \sigma_r, f_r) = -\frac{1}{2}|X| \log \sum_{x\in X} \frac{(f_r(I_B(x)) - I_A(x))^2}{\phi(x - r)} + C,    
\end{equation}
where $X$ is the set of all considered spatial locations in the image. 

To derive a global loss $\mathcal{L}$, we normalize and average local $\mathcal{L}_r$ over all $r\in X$. Although this no longer corresponds to a proper log-likelihood, this nevertheless yields a meaningful global loss that is efficient to optimize. Different modeling assumptions for $f_r$ lead to different loss functions \cite{roche2000unifying}. Not assuming a particular form for $f$ leads to the estimation of conditional expectation (Eq. \ref{eq:correlation_ratio_conditional_expectation}), used by RaPTOR and SRWCR. Local cross-correlation (for intra-modality registration) can be derived from assuming affine form \cite{roche2000unifying,avants2008symmetric}. In the context of rigid registration and global CR, a polynomial $f$ has been proposed \cite{roche2001rigid}.

To find the optimal compromise in modeling assumptions (strict vs. loose), we propose to model $f$ as a linear combination of learned non-linear basis functions
\begin{equation}
    f_{\theta_r, \omega}(x) := \sum_{j=1}^J \theta^{(j)}_r \psi^{(j)}_{\omega}(x)\label{eq:linear_basis}
\end{equation}
where $\theta^{(j)}_r \in \mathbb{R}$ are the local coefficients, and $\psi_{\omega}$ the global basis functions parametrized by $\omega$. We define $\psi_{\omega}$ as a small fully connected neural network, and the parameters can be learned jointly with the deformation (via gradients) while $\theta_r$ can be solved in closed form. In our experiments, relatively small number of terms $J$ (e.g. 4) is enough for good results (saving compute), and having too high $J$ can be detrimental by allowing for poorly matching images with high likelihood.

Maximizing for coefficients $\theta_r$ is a standard weighted linear least squares with input-output pairs $(I_B(x), I_A(x))_{x\in X}$ and input transformation functions $\psi^{(j)}_{\omega}$, yielding the input matrix $\Psi_{\omega} \in \mathbb{R}^{|X| \times J}$, $\left(\Psi_{\omega}\right)_{i, j} := \psi^{(j)}_{\omega}(I_B(x_i))$, the output vector, $y \in \mathbb{R}^{|X|}$, $y_i := I_A(x_i)$, and the diagonal weight matrix $W_r \in \mathbb{R}^{|X|\times |X|}$, $\left(W_r\right)_{i,i} := \frac{1}{\phi(x_i - r)}$. The well-known solution is:
$$
\widehat{\theta}_r:=\underset{\theta_r}{\operatorname{argmax}}\ \mathcal{L}_r = \left(\Psi_{\omega}^TW_r\Psi_{\omega}\right)^{-1}\Psi_{\omega}^T W_r y.
$$

As the final loss we average the formula $\max_{\theta_r} \mathcal{L}_r$ over all $r\in X$ while removing the additive and multiplicative constants:
\begin{equation}\label{eq:final_loss}
    \mathcal{L} := \frac{1}{|X|}\sum_{r\in X} \log \sum_{x\in X} \frac{(f_{\widehat{\theta}_r, \omega}(I_B(x)) - I_A(x))^2}{\phi(x - r)}.
\end{equation}
To use the loss to learn the deformation $d$ one has to simply substitute $I_B$ with $I_B \circ d$. We use average over the log-likelihood instead of likelihood followed by local normalization (as with CR) since the log-form has the same scale invariance property (with respect to the derivatives) and avoids computing the local variance (see the ablation study in Section \ref{sec:ablation_study}).
Computing Eq. \ref{eq:final_loss} naively is expensive. However, assuming that the considered locations $X$ lie on a regular grid, substituting $f_{\theta_r, \omega}$ in Eq. \ref{eq:linear_basis} into Eq. \ref{eq:final_loss}, and rearranging terms yields a formulation of the loss $\mathcal{L}$ that can be calculated efficiently with convolutions:

\begin{equation}
\begin{aligned}
\mathcal{L}
= \frac{1}{|X|} \sum_{r\in X} \log \biggl[\sum_{j=1}^J \sum_{k=1}^J \widehat{\theta}^{(j)}_r \widehat{\theta}^{(k)}_r \left((\psi^{(j)}_{\omega} \circ I_B) (\psi^{(k)}_{\omega} \circ I_B) \ast \frac{1}{\phi}\right)(&r)\\
- 2 \sum_{j=1}^J \widehat{\theta}^{(j)}_r \left((\psi^{(j)}_{\omega} \circ I_B) I_A \ast \frac{1}{\phi}\right)(r)
+ \left(I^2_A \ast \frac{1}{\phi}\right)(&r)\biggl].
\end{aligned}
\end{equation}
Similarly, the terms required to calculate the least squares solution $\widehat{\theta}$ can be expressed with convolutions as
\begin{equation}
\begin{cases}
    \left(\Psi_{\omega}^TW_r\Psi_{\omega}\right)_{j,k} &= \left((\psi^{(j)}_{\omega} \circ I_B) (\psi^{(k)}_{\omega} \circ I_B) \ast \frac{1}{\phi}\right)(r)\\
    \left(\Psi_{\omega}^T W_r y\right)_j & = \left((\psi^{(j)}_{\omega} \circ I_B) I_A \ast \frac{1}{\phi}\right)(r).
\end{cases}
\end{equation}
NB. convolution, given any functions $h_1, h_2: \mathbb{R}^n \to \mathbb{R}$, is defined as the sum $(h_1 \ast h_2)(x) := \sum_{x\in X} h_1(x) h_2(x - r)$.

\subsection{Further details}

\textbf{Incorporating derivative information:} The method allows efficient registration of multichannel data. We take advantage of this and augment input images with additional spatial derivative magnitude channel. This slightly improved the results (see the ablation study in Section \ref{sec:ablation_study}). A similar approach was used with the polynomial $f$ for the rigid global CR registration in \cite{roche2001rigid}.
\\\\
\textbf{Practical implementation:} We incorporate the metric into a multi-resolution (see, e.g., \cite{avants2009advanced,modat2010fast}) registration pipeline with bidirectional formulation via scaling and squaring \cite{arsigny2006log}. We apply the similarity loss in both directions and take the average. We use convolutions with stride $3$ to save computational cost, but we shift the sampling grid randomly over iterations, leading to equal weighting for all spatial locations. We define the kernel $\frac{1}{\phi}$ as the Gaussian kernel (radially symmetric and separable) and truncate it at $3$ standard deviations. The method is implemented in PyTorch and the optimization is performed with gradients of loss w.r.t. the deformation $d$ and the neural network parameters $\phi$ via automated differentation and Adam \cite{kingma2014adam} optimizer. For regularization of the deformation we use the bending energy penalty \cite{rueckert1999nonrigid}. In the experiments, one registration took under $2$ minutes on a modern GPU (V100), and under $1$ minute with $2$ GPUs. The evaluation of similarity and its derivative took (combined time) for the highest resolution level approximately from $50\si{\milli\second}$ to $500\si{\milli\second}$ (volume size differed a lot between the datasets). The method is available at \href{https://github.com/honkamj/locor}{https://github.com/honkamj/locor}.

\section{Experiments}

We conducted experiments on three datasets: two real and one semi-synthetic. The code for all experiments is available at \href{https://github.com/honkamj/locor-experiments}{https://github.com/honkamj/locor-experiments}. 

We evaluated \textbf{abdomen MRI-CT} registration on Learn2Reg 2021 \cite{clark2013cancer,Akin2016-rh,Erickson2016-vh,Linehan2016-et,hering2022learn2reg} data (CC BY 3.0 license) originally from The Cancer Imaging Archive (TCIA) project, which contains 8 sets of MRI-CT image pairs with evaluation based on anatomical segmentation masks of abdominal organs (dice score). We omitted one subject (TCIA 0006) from the test set due to a very different appearance compared to the other images, rendering our hyperparameter optimization setup meaningless.

We used CERMEP-IDB-MRXFDG database \cite{merida2021cermep} (CCO license) with $37$ subjects for evaluating \textbf{head MRI-CT} registration. Since the database contains no labels for evaluation, we generated pseudo-CT images using a deep learning image-to-image translation method designed for geometrically accurate cross-modality synthesis \cite{honkamaa2023deformation}. The mean absolute error (MAE) between the pseudo-CT and registered CT images was used for evaluation. Based on our qualitative visual analysis, the metric corresponds well to the actual registration performance.

\begin{table}[t]
\centering
\setlength{\tabcolsep}{2.4pt}
\caption{Results of the main experiment showing mean and standard deviation over the test cases of each dataset. The result for each test case was computed as the median over 5 test runs (see Section \ref{sec:main_study}). $N_b/N$ refers to the ratio of test subjects for which the method beats our method, averaged over the test runs. corrField was not tested on Head MRI-CT since the implementation could not handle images with different coordinate systems.}\label{tab1}
\begin{tabular}{llclclc}
\toprule
\multicolumn{1}{c}{} & \multicolumn{2}{c}{\scriptsize{Abdomen (MRI-CT)}} & \multicolumn{2}{c}{\scriptsize{Head (MRI-CT)}} & \multicolumn{2}{c}{\scriptsize{Head (MRI T2-PD)}}\\
\cmidrule(r){2-3} \cmidrule(r){4-5} \cmidrule(r){6-7}
Method & \multicolumn{1}{c}{Dice $\uparrow$} & $N_{b}/N$ & \multicolumn{1}{c}{MAE $\downarrow$} & $N_{b}/N$ & \multicolumn{1}{c}{TRE $\downarrow$} & $N_{b}/N$\\
\midrule
       NiftyReg (MIND) & $0.68 (0.30)$ & $0.0/4$ & $111.78 (8.55)$        & $0.0/34$ & $0.52 (0.09)$        & $0.0/49$  \\
        NiftyReg (NMI) & $0.67 (0.31)$ & $0.0/4$ & $77.80 (17.38)$        & $6.2/34$ & $0.96 (0.29)$        & $0.0/49$  \\
      ANTs (Mattes MI) & $0.66 (0.32)$ & $0.0/4$ & $85.36 (12.14)$        & $0.4/34$ & $1.19 (0.30)$        & $0.0/49$  \\
             corrField & $0.80 (0.10)$ & $0.0/4$ & -                      & -        & $1.62 (0.66)$        & $0.0/49$  \\
     SRWCR$^{\dagger}$ & $0.89 (0.03)$ & $1.2/4$ & $76.89 (10.65)$        & $1.6/34$ & $0.30 (0.06)$        & $0.0/49$  \\
   MINDSSC$^{\dagger}$ & $0.90 (0.03)$ & $1.0/4$ & $77.47 (11.82)$        & $3.8/34$ & $0.56 (0.36)$        & $0.0/49$  \\
        MI$^{\dagger}$ & $0.85 (0.04)$ & $0.0/4$ & $81.31 (13.01)$        & $2.8/34$ & $0.56 (0.12)$        & $0.0/49$  \\
 Locor polynom. (ours) & $0.85 (0.07)$ & $0.2/4$ & $79.64 (19.74)$        & $1.0/34$ & $0.20 (0.18)$        & $11.4/49$ \\
 \midrule Locor (ours) & $0.90 (0.02)$ &         & $\bm{65.18} (11.91)^*$ &          & $\bm{0.17} (0.02)^*$ &           \\
\bottomrule
\multicolumn{7}{l}{$^*$ Wilcoxon signed-rank test has p-value $<0.005$ compared to all the baselines.}\\
\multicolumn{7}{l}{$^{\dagger}$ Our reimplementation to a setup similar with our method.}
\end{tabular}
\label{table:results}
\end{table}

We generated a semi-synthetic dataset for \textbf{head MRI T2-PD} registration from IXI\footnote{\href{http://brain-development.org/ixi-dataset/}{http://brain-development.org/ixi-dataset/}} dataset (CC BY-SA 3.0 license) by deforming PD images with random synthetic deformations, consisting of rigid and elastic components; the latter were generated by scaling and squaring \cite{arsigny2006log} from Gaussian smoothed white noise \cite{arsigny2006log}. The target registration error (TRE) in the voxels with respect to the known deformation was used as an evaluation metric (with background masked out). We used a randomly chosen subset of 52 subjects.

\subsection{Hyperparameter optimization}\label{sec:hyperparameter_optimization}

For all datasets, we set aside three subjects as a validation set. For each method (including ours), we ran 100 trials on the validation set to optimise hyperparameters using the Heteroscedastic Evolutionary Bayesian Optimisation (HEBO) sampler \cite{cowen2022hebo}. HEBO can model heteroscedastic noise, which is relevant here, as the observed noise is expected to vary substantially across the hyperparameter space. After this initial search, we fitted HEBO’s Gaussian process model to the sampled parameter–performance pairs and selected the five most promising hyperparameter candidates. For each of these, we carried out five repeated runs on the validation set and selected the configuration with the highest mean performance. This final selection procedure yielded the hyperparameters used in our reported results. The complete setup is documented in the accompanying codebase.

\subsection{Comparison to earlier methods}\label{sec:main_study}

As baselines we used well-establised methods NiftyReg \cite{rueckert1999nonrigid,ourselin2001reconstructing,modat2010fast,modat2014global} (with both MI and MIND \cite{heinrich2012mind}) and Advance Normalization Tools (ANTs) \cite{avants2009advanced}, as well as corrfield\cite{heinrich2015estimating,hansen2021graphregnet} which did well on Abdomen MRI-CT in Learn2reg 2021 \cite{hering2022learn2reg}. We implemented SRWCR \cite{gong2018nonrigid} into identical multi-resolution setup as our method, and for fair comparison, we did that also for mutual information via Parzen windowing (implementation from \cite{guo2019multi}) and MINDSSC \cite{heinrich2013towards,chen2022transmorph} (later variant of MIND). We also compared against our method replaced with the fixed polynomial basis (Locor polynom.).

The results of the main experiment are given in Table \ref{table:results}. Although the hyperparameter optimization setup outlined in Section \ref{sec:hyperparameter_optimization} aimed to obtain robust hyperparameters, several baselines still exhibited occasional instability on the test set. While this may partly reflect genuine inherent instability, we also suspect that the optimisation procedure may still favour non-robust configurations (e.g. unnecessarily large learning rates). To mitigate this, we report test-set performance as the median over five repeated runs per test case. For our method, the mean and median performance are very close. Also, since the metrics used for Head MRI-CT and MRI T2-PD datasets had no upper bound, we clipped them to $120.0$ Hounsfield units and $3.0$ voxels respectively to reduce the influence of extreme failure cases.

Our method outperformed the baselines in head MRI-CT and head T2-PD registration, while in the abdominal MRI-CT registration our method performed similarly to the best baselines, although the dataset is too small to draw statistically significant conclusions.

\subsection{Ablation study}\label{sec:ablation_study}

\begin{table}[t]
\centering
\caption{Results of the ablation study showing mean and standard deviation over the validation cases of each dataset. The result for each validation case was computed as mean over 5 runs. \textbf{Log vs. Normalized}: Whether to use the logarithmic form (Eq. \ref{eq:final_loss}) or to exponentiate and normalize the local log-likelihoods.
\textbf{Learned vs. Polynom.}: Whether to learn $g$ or to use the fixed polynomial basis.
\textbf{Deriv.}: Augment input volumes with derivative magnitude channel.
}
\begin{tabular}{lcccccc}
\toprule

 & \scriptsize{Abdomen (MRI-CT)} & \scriptsize{Head (MRI-CT)} & \scriptsize{Head (MRI T2-PD)}\\
 \cmidrule(r){2-2} \cmidrule(r){3-3} \cmidrule(r){4-4}
 & Dice $\uparrow$ & MAE $\downarrow$ & TRE $\downarrow$\\
        \midrule Log + Learned & $0.904$ ($0.005$) & $59.7$ ($10.8$)   & $0.160$ ($0.016$)    \\
        Log + Learned + Deriv. & $0.905$ ($0.006$) & $58.2$ ($9.5$)    & $0.163$ ($0.013$)    \\
       \midrule Log + Polynom. & $0.901$ ($0.003$) & $68.1$ ($10.7$)   & $0.162$ ($0.016$)    \\
       Log + Polynom. + Deriv. & $0.900$ ($0.006$) & $64.8$ ($10.4$)   & $0.172$ ($0.015$)    \\
 \midrule Normalized + Learned & $0.898$ ($0.009$) & $62.4$ ($11.2$)   & $0.184$ ($0.013$)    \\
 Normalized + Learned + Deriv. & $0.903$ ($0.006$) & $59.1$ ($8.1$)    & $0.178$ ($0.010$)    \\
     %\midrule Parzen windowing & $0.896$ ($0.008$) & $61.0$ ($12.2$)                & $0.227$ ($0.019$)    \\
\bottomrule
\end{tabular}
\label{table:ablation_study_results}
\end{table}

Before the main experiment, we performed an ablation study on the validation set on different variants of the algorithm. The results are shown and the variants are explained in Table \ref{table:ablation_study_results}. For polynomial basis we searched degrees $\leq 5$ and $\leq 3$ for variants without and with derivative magnitude channel, respectively (maximum of $6$ and $10$ terms, respectively).

In conclusion, using the learned $g$ systematically improved the results on the validation set compared to using the fixed polynomial basis. Further evidence was provided by including the polynomial basis variant in the main experiment on the test set ("Locor polynom." in Table \ref{table:results}), where learning the basis improved the results on all three datasets. Including the derivative magnitude channel slightly improved the results in the MRI-CT tasks but not in the MRI T2-PD registration, and we did include it for the test set experiments on that dataset. The logarithmic variant performed slightly better than the normalized one, making it a trivial choice due to cheaper computation.

\section{Discussion}

We provided further evidence that local functional dependence is a good similarity measure for generic multimodal registration. Modeling the function parametrically as a linear combination of learned basis functions further improved the performance, and the method outperformed all the baselines on two out of three datasets. The main downside of the method is the additional complexity due to the learning of the small neural network in conjunction, making optimization dynamics more difficult to understand or predict.

\section*{Acknowledgments}

This work was supported by the Academy of Finland (Flagship programme: Finnish Center for Artificial Intelligence FCAI, and grants 336033, 352986) and EU (H2020 grant 101016775 and NextGenerationEU). We also acknowledge the computational resources provided by the Aalto Science-IT Project.

\bibliographystyle{abbrv}
\bibliography{references}

\end{document}